\documentclass{article}
\usepackage[final,nonatbib]{neurips_2024}

\makeatletter
\renewcommand{\@noticestring}{}  
\makeatother

\usepackage[utf8]{inputenc} 
\usepackage[T1]{fontenc}    
\usepackage[hidelinks]{hyperref}
\usepackage{url}            
\usepackage{booktabs}       
\usepackage{amsfonts}       
\usepackage{nicefrac}       
\usepackage{microtype}      
\usepackage{xcolor}         
\usepackage[frozencache,cachedir=.]{minted}         
\usepackage{xcolor}
\usepackage{graphicx}       

\usepackage[backend=biber,style=numeric]{biblatex}
\makeatletter

\def\NewStructureName#1{}
\def\AssignStructureRole#1#2{}
\def\NewTaggingSocket#1#2{}
\def\NewTaggingSocketPlug#1#2#3{}
\def\AssignTaggingSocketPlug#1#2{}
\def\UseStructureName#1{}
\def\tagstructbegin#1{}

\makeatother

\usepackage{tcolorbox}
\tcbuselibrary{skins,breakable}

\definecolor{OrcUserColor}{RGB}{60, 60, 60}
\definecolor{OrcAgentColor}{RGB}{120, 120, 255}
\definecolor{OrcToolColor}{RGB}{200, 100, 0}
\definecolor{OrcFailedColor}{RGB}{220, 20, 60}

\newtcolorbox{orchestralusermessage}{
  enhanced,
  breakable,
  colback=white,
  colframe=OrcUserColor,
  boxrule=0.6pt,
  arc=1.5mm,
  width=\linewidth,
  top=8pt,
  bottom=8pt,
  left=10pt,
  right=10pt,
  fontupper=\small,
  attach boxed title to top left={yshift=-6pt, xshift=3.5mm},
  boxed title style={colback=white, colframe=white, boxrule=0pt},
  coltitle=OrcUserColor,
  fonttitle=\sffamily,
  title={\raisebox{0pt}[0pt][0pt]{User}},
}

\newtcolorbox{orchestralagentmessage}{
  enhanced,
  breakable,
  colback=white,
  colframe=OrcAgentColor,
  boxrule=0.6pt,
  arc=1.5mm,
  width=\linewidth,
  top=8pt,
  bottom=8pt,
  left=10pt,
  right=10pt,
  fontupper=\small,
  attach boxed title to top left={yshift=-6pt, xshift=3.5mm},
  boxed title style={colback=white, colframe=white, boxrule=0pt},
  coltitle=OrcAgentColor,
  fonttitle=\sffamily,
  title={\raisebox{0pt}[0pt][0pt]{Agent}},
}

\newtcolorbox{orchestraltoolmessage}[1]{
  enhanced,
  breakable,
  colback=white,
  colframe=OrcToolColor,
  boxrule=0.5pt,
  arc=1mm,
  width=\linewidth,
  left=1.5em,
  right=1em,
  top=6pt,
  bottom=6pt,
  fontupper=\small,
  attach boxed title to top left={yshift=-5pt, xshift=3.5mm},
  boxed title style={colback=white, colframe=white, boxrule=0pt},
  coltitle=OrcToolColor,
  fonttitle=\sffamily\small,
  title={\raisebox{0pt}[0pt][0pt]{#1}},
}

\newtcolorbox{orchestraltoolerrormessage}[1]{
  enhanced,
  breakable,
  colback=white,
  colframe=OrcFailedColor,
  boxrule=0.6pt,
  arc=1mm,
  width=\linewidth,
  left=1.5em,
  right=1em,
  top=6pt,
  bottom=6pt,
  fontupper=\small,
  attach boxed title to top left={yshift=-5pt, xshift=3.5mm},
  boxed title style={colback=white, colframe=white, boxrule=0pt},
  coltitle=OrcFailedColor,
  fonttitle=\sffamily\small,
  title={\raisebox{0pt}[0pt][0pt]{#1}},
}

\setminted{
    style=vs,                 
    fontsize=\small,
    frame=lines,
    bgcolor=gray!5,
}

\title{Orchestral AI: A Framework for Agent Orchestration}

\author{%
  Alexander Roman\\
  Orchestral AI\\
  \texttt{alex@orchestral-ai.com} \\
  \And
  Jacob Roman \\
  Orchestral AI \\
  \texttt{jake@orchestral-ai.com} \\
}

\begin{document}

\maketitle

\begin{abstract}
The rapid proliferation of LLM agent frameworks has forced developers to choose between vendor lock-in through provider-specific SDKs and complex multi-package ecosystems that obscure control flow and hinder reproducibility. Integrating tool calling across multiple LLM providers remains a core engineering challenge due to fragmented APIs, incompatible message formats, and inconsistent streaming and tool-calling behavior, making it difficult to build portable, reliable agent systems.
We introduce Orchestral, a lightweight Python framework that provides a unified, type-safe interface for building LLM agents across major providers while preserving the simplicity required for scientific computing and production deployment. Orchestral defines a single universal representation for messages, tools, and LLM usage that operates seamlessly across providers, eliminating manual format translation and reducing framework-induced complexity.
Automatic tool schema generation from Python type hints removes the need for handwritten descriptors while maintaining type safety across provider boundaries. A synchronous execution model with streaming support enables deterministic behavior, straightforward debugging, and real-time interaction without introducing server dependencies. The framework's modular architecture cleanly separates provider integration, tool execution, conversation orchestration, and user-facing interfaces, enabling extensibility without architectural entanglement.
Orchestral supports advanced agent capabilities found in larger frameworks, including rich tool calling, context compaction, workspace sandboxing, user approval workflows, sub-agents, memory management, and MCP integration. Additional mechanisms such as context protection, advanced tool-context tracking, cost limits, and easy LaTeX integration support safe, auditable, and cost-aware agent operation. An optional user interface can be easily added to any agent application.
Orchestral’s modular architecture enables a single lightweight codebase to support both experimental and production-grade LLM agents without sacrificing debuggability, extensibility, or deployment simplicity.
\end{abstract}

\section{Introduction}

Large Language Models (LLMs) have demonstrated remarkable capabilities when augmented with tool calling (the ability to invoke external functions to retrieve information, execute code, or interact with external systems). However, integrating tool calling across multiple LLM providers remains a significant engineering challenge. Each provider exposes different APIs, requires different message formats, handles tool schemas differently, and exhibits unique streaming behaviors. This fragmentation forces developers to choose between vendor lock-in and maintaining provider-specific code paths.

The proliferation of AI agent frameworks has attempted to address this complexity, but existing solutions present significant tradeoffs. Enterprise frameworks like LangChain~\cite{langchain} offer extensive integrations but impose complex multi-package architectures that obscure control flow. Multi-agent frameworks like CrewAI~\cite{crewai} provide role-based orchestration but struggle with debugging and production deployment. Low-code platforms like N8N~\cite{n8n} prioritize visual workflow building at the expense of programmatic control. Provider-specific SDKs like Claude Agent SDK~\cite{claude-sdk} offer deep integration but lock users into a single vendor and offer little visibility or customizability into the underlying agent architecture.

For scientific computing researchers, these frameworks present particular challenges. Reproducibility demands understanding exactly what code executes and when. Long-running computations require robust error handling and interruption support. Exploratory analysis benefits from streaming feedback. Publication workflows need documentation and export capabilities. Yet most frameworks prioritize either enterprise deployment scenarios or impose architectural complexity that hinders research iteration.

\textbf{Orchestral} addresses these challenges through architectural simplicity. The framework provides a unified, type-safe interface across multiple LLM providers while maintaining a synchronous architecture that makes control flow explicit and debugging straightforward. Automatic schema generation from Python type hints enables trivial tool definition while ensuring type safety across provider format conversions. Clear separation of concerns with the Agent as central orchestrator enables extensibility without complexity. The result is a framework suitable for both production deployment and research exploration, implemented in a single lightweight Python codebase.

This paper describes Orchestral's architecture, key features, and design philosophy. We position the framework in the current landscape of agent platforms, explain technical decisions including the choice of synchronous execution, and demonstrate features particularly relevant to researchers including LaTeX export, cost tracking, and reproducible workflows.

\textbf{Getting Started.} Orchestral is available for installation via pip:

\begin{minted}[fontsize=\small, bgcolor=gray!5]{bash}
pip install orchestral-ai
\end{minted}

To get started, refer to the quickstart guide in the \href{https://github.com/orchestralAI/orchestral-ai.git}{\texttt{orchestral-ai} repository}. For comprehensive documentation, tutorials, and API reference, visit \url{https://orchestral-ai.com}.

\section{Related Work and Positioning}

\subsection{The AI Agent Framework Landscape}

The rapid advancement of LLM capabilities has spawned numerous frameworks for building agent systems. We categorize these into five primary approaches:

\textbf{General-Purpose Orchestration Frameworks.} LangChain~\cite{langchain} represents the dominant approach: a layered, modular architecture built on chain-based sequential processing. LangChain provides extensive integrations (hundreds of providers and tools) through multiple packages (\texttt{langchain-core}, \texttt{langchain-community}, integration-specific packages). While this modularity enables flexibility, it introduces complexity. Developers must navigate multiple abstraction layers, understand the LangGraph state machine for complex workflows, and manage dependencies across packages. The framework excels at predictable linear workflows and retrieval-augmented generation but requires significant expertise to debug.

\textbf{Autonomous Agent Frameworks.} AutoGPT~\cite{autogpt} pioneered fully autonomous agents with a three-tier task management system: task creation agents that parse high-level goals, task prioritization agents that order operations, and task execution agents that perform work~\cite{autogpt-arch}. This architecture enables impressive autonomy but introduces significant complexity: self-prompting chains can be difficult to trace, debugging multi-agent coordination requires understanding inter-agent communication, and the framework's power comes at the cost of predictability. The split between backend (FastAPI, PostgreSQL) and frontend (Next.js) creates a heavy deployment footprint unsuitable for simple use cases.

\textbf{Multi-Agent Systems.} CrewAI~\cite{crewai} focuses on orchestrating multiple role-playing agents that collaborate on complex tasks. Its role-based architecture simplifies multi-agent coordination but reveals significant limitations: poor logging infrastructure makes debugging challenging, customization is constrained, and transitioning to production requires implementing monitoring and error recovery mechanisms independently. Users report particular difficulty with open-source models and coordination complexity as agent count scales~\cite{crewai-limitations}.

\textbf{Workflow Automation Platforms.} N8N~\cite{n8n} approaches AI agents through visual workflow building, targeting business process automation rather than programmatic agent development. Its drag-and-drop interface prioritizes accessibility over control, making it suitable for no-code users but limiting for researchers who require programmatic access, fine-grained control, and integration with scientific computing tools.

\textbf{Provider-Specific SDKs.} Anthropic's Claude Agent SDK~\cite{claude-sdk} provides deep integration with Claude, including automatic context compaction, MCP extensibility, and specialized features like subagents and hooks. The SDK powers Claude Code, an IDE integration requiring Electron runtimes and VSCode/JetBrains infrastructure. While this tight integration enables sophisticated coding assistance, it creates significant deployment constraints: agents become tightly coupled to development environments, deployment requires IDE infrastructure, and embedding agents in research projects or production applications requires working around the IDE-centric architecture. The commitment to a single provider eliminates the ability to compare models or optimize for cost across vendors.

\subsection{Orchestral's Differentiation}

Orchestral occupies a unique position through several key architectural decisions:

\begin{itemize}
    \item \textbf{Provider agnosticism without complexity:} Unlike LangChain's multi-package ecosystem or Claude SDK's single-provider focus, Orchestral provides a unified interface across all major providers through a clean abstraction layer. Write once, run anywhere.

    \item \textbf{Deterministic execution for reproducibility:} Orchestral uses synchronous execution, ensuring that agent runs are deterministic and reproducible. This enables straightforward debugging with complete stack traces and makes research workflows more reliable. Streaming support is implemented via generators, providing real-time feedback without event loop non-determinism.

    \item \textbf{Type-safe tool generation:} Automatic schema generation from Python type hints eliminates manual descriptor writing while ensuring type safety across provider format conversions. We provide a simple python decorator \texttt{@define\_tool()} that turns any well written python function into a tool with an input schema, validation, and error handling that can be used with any LLM.

    \item \textbf{Lightweight deployment:} Single Python package, minimal dependencies, embeddable in any project. Unlike IDE-centric frameworks (Claude Code) or server-heavy systems (AutoGPT), Orchestral deploys anywhere Python runs: serverless functions, budget VPS instances, Docker containers, or embedded in research projects. The deployment footprint is pip and nothing more.

    \item \textbf{Production features and research affordances:} Conversation persistence, cost tracking, multi-layered security, and streaming support combined with LaTeX integration, reproducible workflows, and read-before-edit safety mechanisms.

    \item \textbf{Modular by design:} Clear separation of concerns with the Agent as the central orchestrator. The core framework is independent from application-specific code, enabling reuse across contexts (CLI, web, notebooks, APIs).
\end{itemize}

The result is a framework that researchers can understand completely (improving reproducibility), developers can debug straightforwardly (reducing development time), and both can deploy confidently, whether embedding in research projects or deploying to production infrastructure.

\section{Architecture}

\begin{figure}[h]
    \centering
    \includegraphics[width=0.7\textwidth]{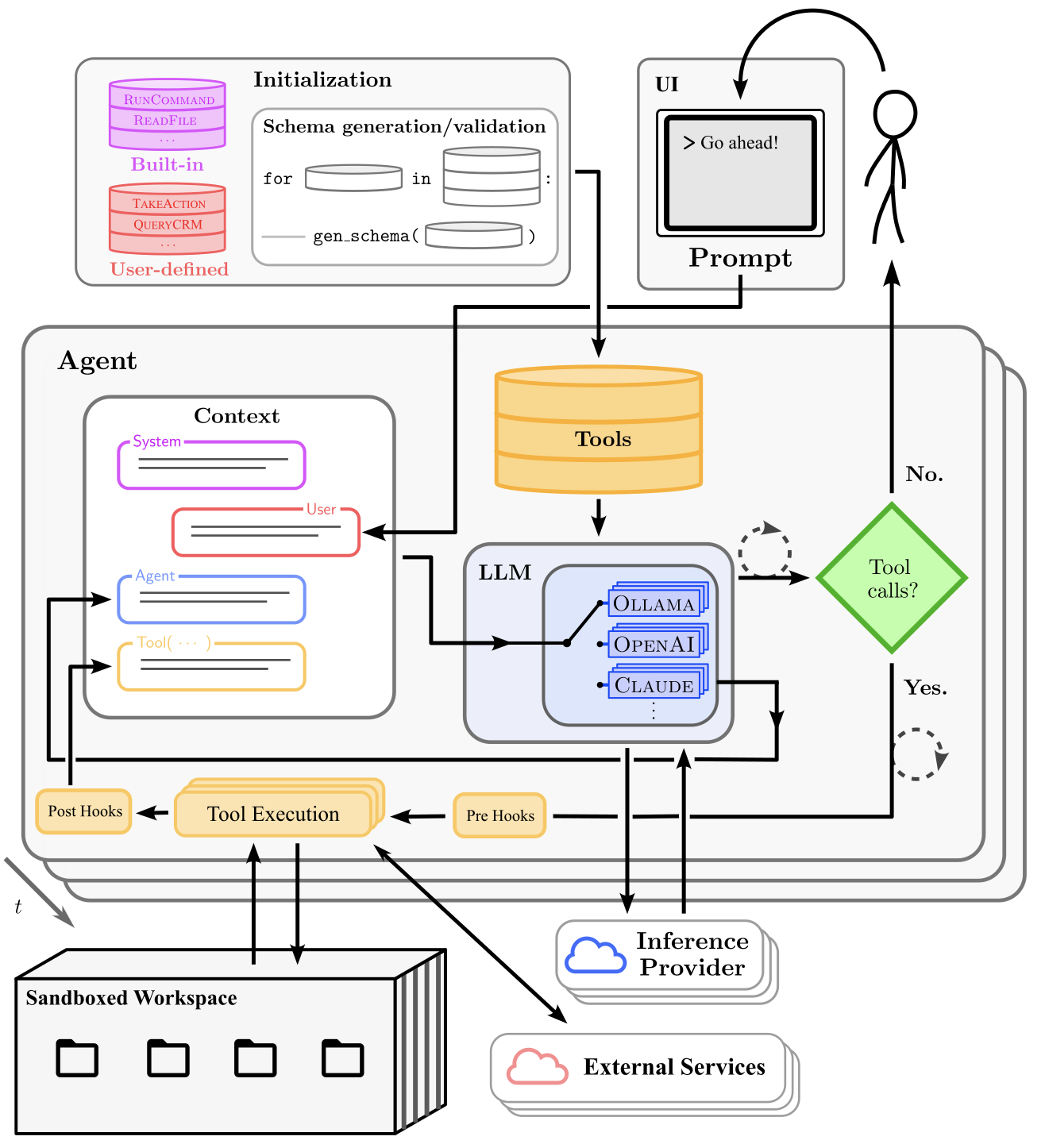}
    \caption{Orchestral architecture centered on the Agent object, which contains the LLM, Tools, and Context. The Agent manages tool execution flow through pre/post hooks, makes tool calling decisions, and updates the conversation context. External components include the sandboxed workspace, UI, and external services.}
    \label{fig:architecture}
\end{figure}

In the Orchestral framework, the Agent object serves as the central orchestrator of conversation and tool execution. Figure~\ref{fig:architecture} shows the complete architecture. The Agent contains three core components: LLM, Tools, and Context. We contend that the combination of these components fundamentally define an Agent. The Agent object also encapsulates the execution flow, tool calling logic, context updates, and more. Each Agent is also given an associated sandboxed workspace where many of its tools operate.
The framework's modular structure separates provider integration, tool execution, conversation management, and user interfaces, enabling extensibility without architectural entanglement.

\subsection{LLM: Provider Abstraction}

The LLM object provides a unified interface across most major providers through an abstract base class. Orchestral's provider support is designed to be easily extensible. At the time of writing, supported providers include OpenAI, Anthropic, Google, Groq, Mistral, AWS Bedrock, and local models via Ollama. Each provider implements a common set of methods:

\begin{itemize}
    \item \textbf{Input Processing:} Convert the framework's unified Context to provider-specific message format
    \item \textbf{API Calls:} Both synchronous and streaming API invocation
    \item \textbf{Output Processing:} Parse provider responses back to unified Response objects
    \item \textbf{Streaming Handling:} Extract text from stream chunks and build final responses
    \item \textbf{Tool Schema Conversion:} Translate tool definitions to provider-specific formats
\end{itemize}

Each provider implements these methods while exposing identical external APIs to application code. From the user's perspective:

\begin{minted}[fontsize=\small, bgcolor=gray!5]{python}
# Switch providers by changing one line
llm = Claude(model='claude-sonnet-4.5')
# llm = GPT(model='gpt-5')
# llm = Gemini(model='gemini-3-pro')

agent = Agent(llm=llm, tools=tools)
# Identical API regardless of provider
\end{minted}

Provider-specific parsers handle message format conversion, tool adapters handle schema conversion (Anthropic's format differs from OpenAI's which in turn differs from Google's), and pricing models enable automatic cost tracking. This design encapsulates all provider complexity within child classes, ensuring application code never imports provider-specific classes directly.

The Ollama provider supports custom host configuration, enabling connection to institutional HPC clusters or private model servers. Many research labs prefer this deployment model for inference privacy, funding independence, and legal compliance reasons. Simply specify the host URL when initializing the Ollama provider to connect to local or institutional infrastructure.

Adding support for a new provider is straightforward: extend the BaseLLM class and implement methods that convert Orchestral's message objects to the provider-specific format, call the API, and parse responses and usage back into the universal Orchestral format.

\subsection{Tools: Type-Safe Tool Framework}

Tools enable LLMs to invoke external functions. The framework provides two approaches for tool definition:

\textbf{Class-based definition} for stateful tools:

\begin{minted}[fontsize=\small, bgcolor=gray!5]{python}
from orchestral.tools.base import BaseTool, RuntimeField

class DataAnalysisTool(BaseTool):
    """Analyze numerical dataset"""

    data_path: str | None = RuntimeField(
        description="Path to CSV data file"
    )
    method: str = RuntimeField(
        default="mean",
        description="Analysis method: mean, median, std"
    )

    def _run(self) -> str:
        import pandas as pd
        df = pd.read_csv(self.data_path)
        result = getattr(df, self.method)()
        return f"Analysis result: {result}"
\end{minted}

\textbf{Decorator-based definition} for stateless functions:

\begin{minted}[fontsize=\small, bgcolor=gray!5]{python}
from orchestral import define_tool

@define_tool()
def calculate_energy(mass: float, c: float = 299792458.0):
    """Calculate relativistic energy E=mc²

    Args:
        mass: Mass in kilograms
        c: Speed of light in m/s (default: exact value)
    Returns:
        Energy in joules
    """
    return mass * c ** 2
\end{minted}

In both cases, the framework automatically:
\begin{enumerate}
    \item Extracts parameter types from Python type hints
    \item Generates JSON schemas compatible with each provider
    \item Validates inputs via Pydantic
    \item Converts between provider-specific formats
    \item Handles execution and error formatting
\end{enumerate}

\textbf{RuntimeField} vs. \textbf{StateField} enables a critical distinction: RuntimeFields are parameters the LLM provides (appearing in tool schemas), while StateFields maintain internal state invisible to the LLM. An example of the power of this is the persisent terminal tool described in Section~\ref{sec:persistent-sessions}.
This separation enables tools to maintain context across calls without exposing implementation details.

It is simple to turn external functions into tools, they can simply be wrapped with the \texttt{@define\_tool()} decorator:

\begin{minted}[fontsize=\small, bgcolor=gray!5]{python}
from orchestral import define_tool
from my_database import query_database

@define_tool()
def QueryDatabaseTool(query: str, limit: int = 10) -> str:
    """
    <description of the function and parameters>
    """
    results = query_database(query, limit=limit)

    return f'Results:\n{format_results(results)}'
\end{minted}

\subsection{Hooks: Tool Execution Interception}

Hooks intercept tool execution at two points: before execution (pre-hooks) and after (post-hooks). Pre-hooks can approve, reject, or modify tool calls, with the first rejection short-circuiting execution. Post-hooks transform outputs (truncation, summarization, formatting).

Hooks are optional, interchangeable, and customizable. The framework includes several built-in hooks:

\textbf{Pre-execution hooks} (security and approval):
\begin{itemize}
    \item \texttt{SafeguardHook}: LLM-based safety analysis of commands
    \item \texttt{UserApprovalHook}: Three-tier classification (SAFE/APPROVE/UNSAFE)
    \item \texttt{DangerousCommandHook}: Pattern-based blocking of dangerous operations
\end{itemize}

\textbf{Post-execution hooks} (output processing):
\begin{itemize}
    \item \texttt{TruncateHook}: Limit output length to preserve context window
    \item \texttt{SummarizeHook}: LLM-based summarization of large outputs
\end{itemize}

Custom hooks extend a simple interface to implement domain-specific workflows like cost control, logging, or testing. New custom hooks can be written by extending the BaseHook class and implementing the pre/post execution methods.

\subsection{Context: Message History \& Validation}

The Context object manages conversation history, validates message sequences, and enables persistence:

\begin{minted}[fontsize=\small, bgcolor=gray!5]{python}
class Context:
    messages: list[Message]
    total_cost: float
    metadata: dict

    def add_message(self, message: Message):
        """Add and validate message"""

    def remove_orphaned_tool_results(self):
        """Clean unmatched tool results"""

    def save_json(self, path: str):
        """Serialize entire conversation"""
\end{minted}

Key innovations include:

\textbf{Tool call validation:} LLM APIs require tool calls and results to match precisely. Missing a tool result or providing results without calls causes cryptic errors. Context automatically detects and cleans orphaned results before API calls.

\textbf{Provider-agnostic serialization:} Messages serialize to a unified format independent of which provider generated them. Conversations can be saved, loaded, and continued with a different provider.

\textbf{Cost tracking:} Every API call records token usage and cost. Context aggregates across turns, enabling researchers to monitor computational budgets.

\subsection{Agent: Conversation Orchestration}

The Agent object orchestrates multi-turn conversations with tool calling loops:

\begin{minted}[fontsize=\small, bgcolor=gray!5]{python}
agent = Agent(
    llm=Claude(model='claude-sonnet-4-0'),
    tools=[DataAnalysisTool(), calculate_energy],
    system_prompt="You are a physics research assistant"
)

# Simple text interaction
response = agent.send_text_message("Hello")

# Full tool-calling loop
response = agent.run(
    "Analyze data.csv using median method, then calculate energy for m=1kg",
    max_iterations=8
)

# Streaming with manual tool handling
for chunk in agent.stream_text_message("Explain E=mc²"):
    print(chunk, end='')
\end{minted}

The \texttt{run()} method implements robust multi-turn execution:
\begin{enumerate}
    \item Fix orphaned tool results and missing IDs
    \item Send user message to LLM
    \item Receive response (potentially with tool calls)
    \item Execute all tool calls through hook system
    \item Send tool results back to LLM
    \item Repeat until no tool calls or max iterations
    \item Return final assistant message
\end{enumerate}

The Agent supports interruption through an interrupt flag that is checked during tool execution loops. The web UI leverages this to enable users to halt long-running computations without corrupting conversation state.

\subsection{User Interface}

\begin{figure}[h]
    \centering
    \includegraphics[width=0.99\textwidth]{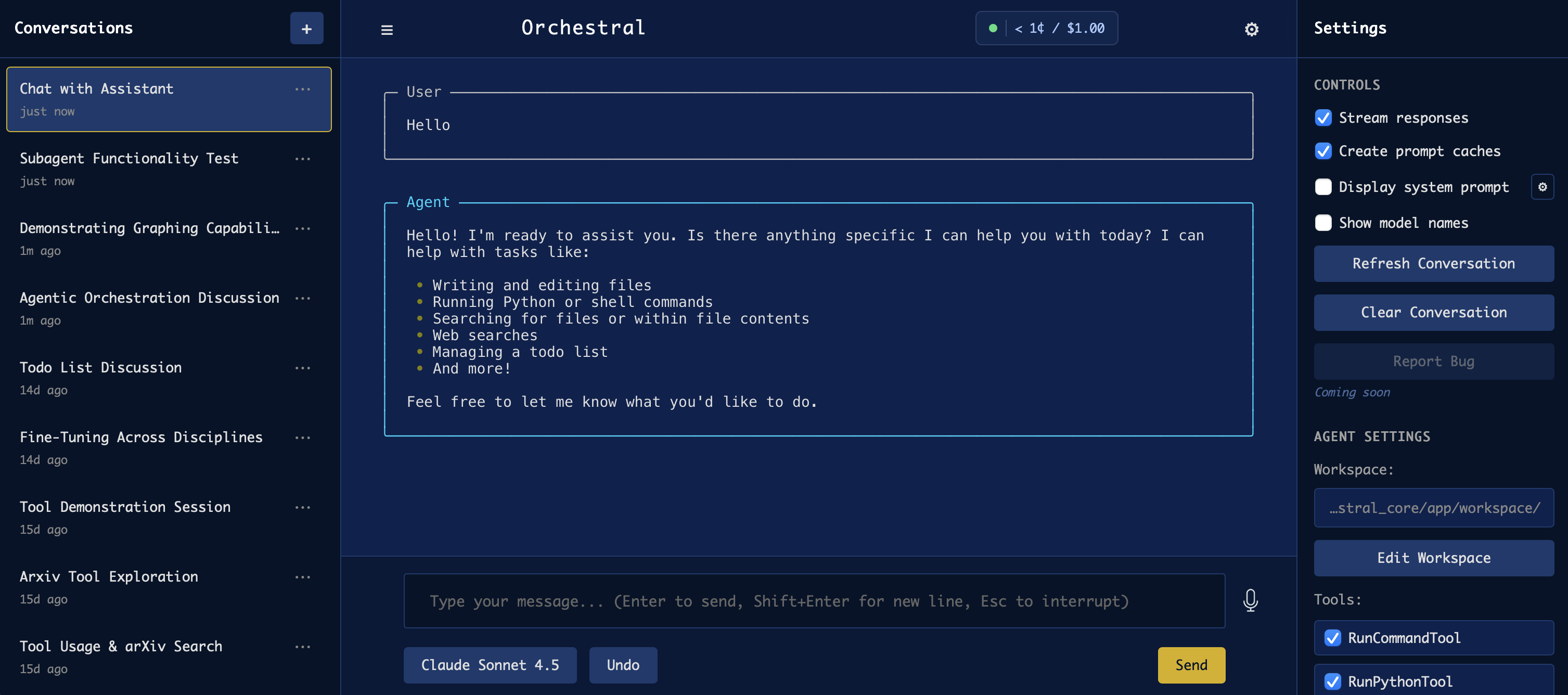}
    \caption{The Orchestral web UI showing a conversation with tool execution and streaming responses.}
    \label{fig:ui-screenshot}
\end{figure}

Orchestral provides a UI which is decoupled from the core Agent, allowing applications to use Orchestral as a library without any interface, or easily add a UI to any Agent objects. Figure~\ref{fig:ui-screenshot} shows a typical session in the interface. The UI is a simple web application that runs locally on the users computer.

In addition to the web UI, the framework includes a command-line interface (CLI) for quick experimentation and debugging. The CLI supports all core Agent features, including tool calling, streaming, and context persistence but does not allow convienent features like interrupts or message editing.

\section{Core Features}

\subsection{Provider Agnosticism}

Orchestral supports all major LLM providers through a unified interface:

\begin{minted}[fontsize=\small, bgcolor=gray!5]{python}
# Switch providers by changing one line
llm = GPT(model='gpt-4')
llm = Claude(model='claude-sonnet-4-0')
llm = Gemini(model='gemini-2.0-flash-exp')
llm = Ollama(model='llama3.1:70b')
llm = Groq(model='llama-3.1-70b-versatile')

agent = Agent(llm=llm, tools=tools)
# Identical API regardless of provider
\end{minted}


Orchestral supports \textit{synthetic LLMs}: classes that abstract provider selection logic behind the standard LLM interface. The \texttt{CheapLLM} router exemplifies this pattern by automatically selecting the cheapest available provider based on which API keys are configured. This abstraction is particularly valuable for applications deployed to users with varying API key configurations, enabling model selection without hardcoding specific providers.

\subsection{Streaming Without Async}

Streaming is implemented through synchronous generators:

\begin{minted}[fontsize=\small, bgcolor=gray!5]{python}
for chunk in agent.stream_text_message(prompt):
    print(chunk, end='', flush=True)
\end{minted}

The framework handles:
\begin{itemize}
    \item Collecting chunks to form completed text
    \item Aggregating interlaced tool call chunks (multiple parallel tools)
    \item Provider-specific usage aggregation (some providers send usage per chunk, others only in final chunk)
    \item Building the final Response object
\end{itemize}

This synchronous streaming eliminates async/await complexity while providing immediate user feedback and enabling interrupts for long-running computations.

\subsection{Multi-Layered Security}

Scientific computing with agents requires protecting against destructive operations. Orchestral provides three security layers through pre-execution hooks:

\textbf{1. Pattern Blocking:}
\begin{minted}[fontsize=\small, bgcolor=gray!5]{python}
DangerousCommandHook()
# Blocks: rm -rf, eval(), exec(), etc.
\end{minted}

\begin{figure}[h]
    \centering
    \includegraphics[width=0.8\textwidth]{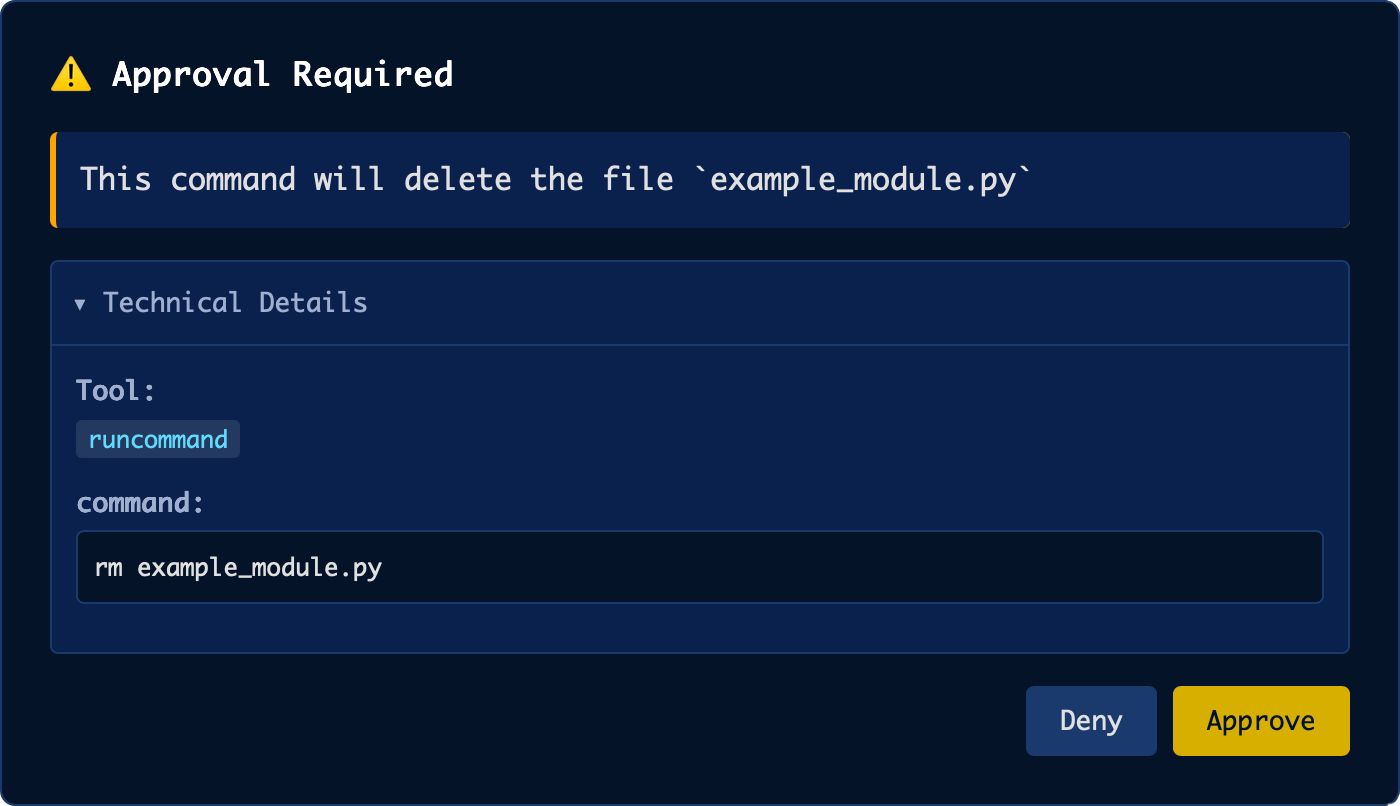}
    \caption{User Approval Hook prompting the user to approve a potentially dangerous unwanted file deletion command.}
    \label{fig:user-approval-hook}
\end{figure}

\textbf{2. User Approval:} The \texttt{UserApprovalHook} implements a three-tier approval system (SAFE: auto-approve, APPROVE: ask user, UNSAFE: reject) as shown in Figure~\ref{fig:user-approval-hook}.

\textbf{3. LLM-as-Judge:}
\begin{minted}[fontsize=\small, bgcolor=gray!5]{python}
SafeguardHook()
# Separate LLM analyzes command safety
\end{minted}

Hooks chain in sequence, with the first rejection short-circuiting execution. Tools also implement safety mechanisms: \texttt{EditFileTool} enforces read-before-edit, preventing blind overwrites.

\subsection{Conversation Persistence \& Cost Tracking}

Every conversation can be serialized to a provider-agnostic JSON format. Conversations saved with Claude can be loaded and continued with GPT-4:

\begin{minted}[fontsize=\small, bgcolor=gray!5]{python}
# Save entire conversation
agent.context.save_json("conversation.json")

# Load and continue with different provider
context = Context.load_json("conversation.json")
agent = Agent(llm=GPT(model='gpt-4'), tools=tools, context=context)
agent.run("Continue where we left off")
\end{minted}

The Context object tracks total cost and token usage across all API calls, enabling researchers to monitor computational budgets in real-time.

\subsection{Cost-Aware Workflows}

Cost-conscious researchers can use different models for different task complexities. The Ollama provider automatically detects local models running on the specified host:

\begin{minted}[fontsize=\small, bgcolor=gray!5]{python}
# Explore with local model (zero cost)
explorer = Agent(llm=Ollama(model='llama3.1:70b'), tools=tools)
explorer.run("Analyze dataset and plan approach")

# Switch to powerful model, reusing context
context = explorer.context  # Preserve conversation
reasoner = Agent(llm=Claude(model='claude-opus-4-0'),
                 tools=tools, context=context)
reasoner.run("Execute the planned analysis")
\end{minted}

This pattern enables budget-conscious iteration: use local models for drafts and refinement, expensive models for final results.

\subsection{Context Integrity \& Validation}

Robust context management prevents common failure modes:

\textbf{Tool call/result matching:} Orphaned tool results (results without corresponding calls) cause LLM API errors. Context automatically detects and removes orphans before each API call.

\textbf{Tool call ID assignment:} Some providers don't assign IDs to tool calls. Framework generates unique IDs to maintain call/result correspondence.

\textbf{Undo and experimentation:} Interactive workflows benefit from undo capability. The Context provides \texttt{undo()} to remove the last user message and all subsequent responses, enabling users to backtrack when exploration goes awry. \texttt{copy()} enables branching conversations, allowing users to try different approaches while preserving the original context.

\begin{minted}[fontsize=\small, bgcolor=gray!5]{python}
# Try an approach
agent.run("Analyze with method A")

# Didn't work, undo and try differently
agent.context.undo()
agent.run("Analyze with method B")

# Or fork the conversation
context_copy = agent.context.copy()
alternative_agent = Agent(llm=llm, context=context_copy)
# Both agents can now proceed independently
\end{minted}

\textbf{Output truncation:} Tool outputs can be enormous (e.g., reading large files). Post-execution hooks truncate or summarize outputs to preserve context for meaningful information:

\begin{minted}[fontsize=\small, bgcolor=gray!5]{python}
agent = Agent(
    llm=llm,
    tools=[ReadFileTool()],
    tool_hooks=[
        TruncateOutputHook(max_chars=5000),
        SummarizeOutputHook(llm=cheap_llm)
    ]
)
\end{minted}


\section{Advanced Capabilities}

\subsection{Composable Safety and Workflow Control}

The hook system enables sophisticated workflow control by composing multiple interception layers. Beyond the security applications discussed earlier, hooks support diverse research and production use cases:

\begin{itemize}
    \item \textbf{Audit trails:} Log all tool executions with parameters and outputs for reproducibility documentation
    \item \textbf{Testing and development:} Replace expensive API calls or long-running computations with mock data during agent development
    \item \textbf{Cost control:} Enforce computational budgets by rejecting operations that would exceed spending limits (see Section~\ref{sec:custom-hooks} for implementation)
    \item \textbf{Experiment tracking:} Record tool usage patterns to analyze agent behavior across runs
    \item \textbf{Progressive automation:} Start with user approval for all operations, gradually whitelist trusted patterns as confidence grows
\end{itemize}

The composability of hooks allows researchers to combine security (SafeguardHook), user control (UserApprovalHook), and efficiency (TruncateOutputHook) in a single agent, creating workflows that balance autonomy with safety.

\subsection{Read-Before-Edit Safety}

The \texttt{EditFileTool} implements transparent safety: it tracks which files have been read in the current conversation. Attempting to edit a file that hasn't been read triggers a rejection with an error message guiding the LLM to read first as shown below:


\begin{orchestraltoolerrormessage}{EditFile: example.txt}
\textbf{Error:} File Not Read

\textbf{Reason:} You must read the file before editing it

\textbf{Context:} Path: example.txt

Use read\_file to see the current content first, then copy the exact text you want to change into old\_string
\end{orchestraltoolerrormessage}

This prevents blind overwrites while remaining completely transparent to the LLM. No special prompting is required; the safety mechanism operates through metadata in the Context layer.

\subsection{External Modification Detection}

Tools also track file hashes to detect external modifications.
This prevents race conditions in collaborative workflows where humans and agents edit simultaneously. Attempting to edit a file that has changed since it was last read triggers a warning, prompting the LLM to re-read the file for current content as shown below:


\begin{orchestraltoolerrormessage}{EditFile: example.txt}
\textbf{Error:} File Modified Since Read

\textbf{Reason:} The file has been modified since you last read it

\textbf{Context:} Path: example.txt

Last Read: 2024-04-08 14:27:12

Modified: 2024-04-08 15:35:23

Use read\_file again to see the current content, then retry your edit with the updated content
\end{orchestraltoolerrormessage}

\subsection{Built-in Tools}

Orchestral provides a curated set of production-ready tools covering common agentic workflows:

\textbf{Filesystem:}
\begin{itemize}
    \item \texttt{ReadFileTool}: Read with encoding detection, line ranges, syntax highlighting
    \item \texttt{WriteFileTool}: Write with automatic backups, directory creation
    \item \texttt{EditFileTool}: Smart editing with modification detection
    \item \texttt{ListDirectoryTool}: Recursive listing with filtering
    \item \texttt{FileSearchTool}: Regex search with context lines
    \item \texttt{FindFilesTool}: Glob-based file finding
\end{itemize}

\textbf{Execution:}
\begin{itemize}
    \item \texttt{RunCommandTool}: Persistent shell (working directory, env vars maintained)
    \item \texttt{RunPythonTool}: Python execution with pre-loaded packages
\end{itemize}

\textbf{Web \& Research:}
\begin{itemize}
    \item \texttt{WebSearchTool}: Search with result extraction
    \item \texttt{ArxivTool}: Academic paper search and retrieval
\end{itemize}

\textbf{Utilities:}
\begin{itemize}
    \item \texttt{TodoRead}/\texttt{TodoWrite}: Task tracking within conversations
    \item \texttt{DisplayImageTool}: Render images in supported interfaces
\end{itemize}

Each tool is production-tested with comprehensive error handling, input validation, and safety mechanisms. The modular design makes adding domain-specific tools straightforward via the \texttt{BaseTool} class or \texttt{@define\_tool()} decorator.

\subsection{Subagents}

Orchestral supports agentic tools: tools that contain their own Agent instance for autonomous task execution. The \texttt{BaseSubagent} class extends \texttt{BaseTool}, enabling tools to perform multi-step reasoning with their own LLM and tool set. Subagents enable complex hierarchical task decomposition while maintaining Orchestral's core principle of explicit, debuggable control flow.

This feature enables hierarchical agentic workflows in which high-level agents focus on strategic orchestration while delegating implementation details to specialized subagents. Subagents are particularly valuable for well-defined tasks requiring exploration and trial-and-error: the subagent sifts through potentially irrelevant content, returning only essential information to the parent agent's context. This improves information density and preserves the parent agent's context window for high-level reasoning. Subagents can recursively invoke other subagents, enabling arbitrary task decomposition.

A representative example is document analysis: extracting structured information from LaTeX source files (which may be organized in diverse ways across projects) benefits from a subagent that explores the file system, reads relevant files, and synthesizes findings. The parent agent receives clean results without context pollution from exploratory file reads. When tasks require repeated execution, specialized system prompts and domain-specific tools can be embedded in subagents, building reusable expertise.

\textbf{Philosophical Note:}
Alfred North Whitehead observed that ``civilization advances by extending the number of important operations which we can perform without thinking about them''~\cite{whitehead1911}. The development of Orchestral AI has been guided by this vision of abstraction.
In a \emph{trivial sense}, the entire framework abstracts away the complexity of LLM provider APIs, tool schema formats, and context management so users can focus on agentic workflows rather than plumbing.
But in a \emph{deeper sense}, features like subagents embody this philosophy: by encapsulating complex reasoning within subagents, high-level agents can operate without micromanaging every detail. This abstraction enables researchers to focus on strategic objectives while delegating operational complexity to specialized components, advancing the state of agentic AI systems.

\subsection{MCP Integration}

The Model Context Protocol (MCP), developed by Anthropic, standardizes tool sharing across AI applications~\cite{claude-sdk}. MCP support enables Orchestral users to:

\begin{itemize}
    \item \textbf{Use MCP servers:} Connect to databases, APIs, and services through standardized MCP server implementations
    \item \textbf{Share tools:} Tools written for Orchestral can be exposed as MCP servers for use by other applications
    \item \textbf{Ecosystem access:} Leverage the growing MCP ecosystem while maintaining Orchestral's type safety and provider agnosticism
\end{itemize}

Integration preserves Orchestral's architecture: MCP tools are adapted to the \texttt{BaseTool} interface, ensuring they work identically across all supported LLM providers. This differs from MCP-only frameworks where tools are tightly coupled to specific providers.

The combination of Orchestral's automatic schema generation and MCP's standardized protocol creates a powerful extensibility model. Tools defined once work across both the Orchestral ecosystem and the broader MCP ecosystem.

\subsection{Custom Hooks}\label{sec:custom-hooks}

Implementing custom hooks enables domain-specific workflows:

\begin{minted}[fontsize=\small, bgcolor=gray!5]{python}
class BudgetControlHook(ToolHook):
    """Reject operations exceeding cost budget"""

    def __init__(self, max_cost: float):
        self.max_cost = max_cost

    def before_call(self, tool, context):
        if context.total_cost > self.max_cost:
            return ToolHookResult(
                approved=False,
                message=f"Budget exceeded: "
                        f"${context.total_cost:.2f} > "
                        f"${self.max_cost:.2f}",
                should_interrupt=True
            )
        return ToolHookResult(approved=True)

agent = Agent(
    llm=expensive_llm,
    tools=tools,
    tool_hooks=[BudgetControlHook(max_cost=10.0)]
)
\end{minted}

\section{Research-Specific Features}

\subsection{LaTeX Integration for Publications}

Scientific workflows often require documenting agent interactions for publication. The web UI includes copy buttons that export conversation snippets as LaTeX code using the provided \texttt{orchestral.tex} module. Researchers add \texttt{\textbackslash{}include\{orchestral.tex\}} to their paper's preamble, which defines custom environments for orchestral messages, then paste formatted conversations into the document body:

\begin{orchestralusermessage}
I just arrived in Paris, do I need my coat?
\end{orchestralusermessage}

\begin{orchestralagentmessage}

\begin{orchestraltoolmessage}{GetWeather( location = "Paris, France" )}
Temperature: 26°C, Clear Skies, Humidity: 42\%, Wind: 10 km/h
\end{orchestraltoolmessage}

The current weather in Paris is 26°C with clear skies, and there's a light breeze at 10 km/h.

You probably won't need your coat, it should be quite pleasant!
\end{orchestralagentmessage}

The module provides colored boxes matching the UI appearance, enabling researchers to maintain reproducible records of agent-assisted analysis integrated directly into papers or theses.

\subsection{Persistent Terminal Sessions}\label{sec:persistent-sessions}

The \texttt{RunCommandTool} supports persistent shell sessions across multiple calls, preserving working directory and environment variables:

\begin{minted}[fontsize=\small, bgcolor=gray!5]{python}
agent.run("cd /data && ls")
agent.run("pwd")  # Returns: /data (cd persisted)

agent.run("export DATA_PATH=/data/experiment1")
agent.run("echo $DATA_PATH")  # Returns: /data/experiment1
\end{minted}

Persistence is optional and controllable via tool parameters, but when enabled, it aligns terminal behavior with how humans naturally use command-line interfaces. This design philosophy—meeting LLMs "where they are" rather than forcing adaptation to framework constraints—reduces cognitive load. By designing tools that match LLM expectations based on their training data (observing human command-line usage), we optimize what we term "LLM-UX": the user experience from the model's perspective. This approach enables complex workflows requiring directory navigation, environment setup, and stateful operations common in scientific computing, while improving reliability for command-line tasks.

\section{Use Cases}

Orchestral has been deployed in multiple scientific research projects, demonstrating its suitability for domain-specific AI agent applications. These applications share common requirements that guided Orchestral's design: reproducible workflows for publication, cost-conscious provider selection, integration with scientific Python ecosystems, and conversation export for documenting research processes. The framework's lightweight deployment model enabled embedding directly in research codebases rather than requiring architectural commitment to agent platforms.

\subsection{HEPTAPOD: High-Energy Physics}

The HEP Toolkit for Agentic Planning, Orchestration, and Deployment~\cite{HEPTAPOD} applies Orchestral to Beyond the Standard Model (BSM) physics. The framework enables agents to interface with domain-specific tools, construct and manage Monte Carlo simulation workflows, and assist in validation pipelines. A representative workflow spans model generation, event simulation, and downstream analysis within a unified, reproducible pipeline. The structured, auditable layer between researchers, LLMs, and computational infrastructure establishes a foundation for transparent, human-in-the-loop systems in particle physics.

\subsection{ASTER: Exoplanet Atmospheres}

The Agentic Science Toolkit for Exoplanet Research~\cite{ASTER} combines tools for downloading planetary parameters from the NASA Exoplanet Archive, generating TauREx forward models, and performing Bayesian retrieval analysis. The agent assists users by proposing alternative modeling approaches, reporting potential issues, and suggesting interpretations. A complete case study of WASP-39b demonstrates ASTER's ability to rapidly execute retrieval analysis, efficiently transitioning between datasets and recovering atmospheric parameters reported in the literature. The unified platform demonstrates how agentic systems can accelerate comparative retrieval studies in exoplanet characterization.

\section{Limitations and Future Work}

\subsection{Current Limitations}

\textbf{Context management:} Automatic context compaction/summarization is not yet implemented. Users must manually manage context when approaching model limits.

\textbf{Parallel tool execution:} Tools execute sequentially within each turn. Some frameworks support parallel tool execution, which could improve performance for independent operations.

\textbf{Multi-agent orchestration:} Orchestral focuses on single-agent workflows. Multi-agent collaboration patterns (like CrewAI's role-based teams) require manual orchestration.

\textbf{Visual/multimodal:} Image understanding is supported (models can analyze images), but image generation and multimodal tool outputs are not yet integrated.

\subsection{Future Directions}

\subsubsection{Multi-Agent Orchestration}

Orchestral's architecture naturally extends to multi-agent scenarios without requiring framework-level coordination primitives. Unlike systems that introduce complex message brokers or specialized agent types, multi-agent patterns emerge from Orchestral's existing abstractions:

\textbf{Hierarchical Coordination:} A manager agent coordinates worker agents through standard tool calls. Each worker is a separate Agent instance with specialized tools and prompts. The manager delegates subtasks by invoking worker agents as tools, with results returning through the normal tool result mechanism. No special inter-agent protocol required, no hidden message passing, just function calls. Debugging is trivial: stack traces show the complete call chain from manager through workers.

\textbf{Agent-to-Agent (A2A) Protocol:} Support for standardized A2A communication will enable agents to request information, delegate tasks, and coordinate work while maintaining Orchestral's type safety and provider agnosticism. Critically, agents communicate via tool calls, not message brokers, preserving the framework's elegant, synchronous architecture. An agent requesting help from another agent simply invokes it as a tool; the callee has no awareness it was invoked by an agent rather than a human.

\textbf{Specialization via Tools:} Multi-agent systems arise from agents with different tool sets and prompts, not from framework-level agent taxonomies. A code-reviewing agent has linting tools; a data-analysis agent has plotting tools; a literature-search agent has web search tools. Specialization is explicit and understandable: inspect the tool list to understand capabilities.

This approach maintains Orchestral's core principle: \textit{no hidden complexity}. Multi-agent coordination is visible in the code, debuggable through standard stack traces, and reproducible through conversation serialization. The same architectural discipline that makes single-agent systems understandable extends to multi-agent scenarios.

\subsubsection{Automatic Context Compaction}

Implement LLM-based summarization of conversation history when approaching context limits. The challenge: preserve critical information while freeing tokens. The solution: treat summarization as a tool that operates on context, maintaining the framework's tool-centric architecture. Compaction strategies (sliding window, semantic compression, hierarchical summarization) become tool implementations, not framework primitives.



\subsubsection{Enhanced MCP Integration}

Deeper integration with the Model Context Protocol ecosystem: host Orchestral tools as MCP servers for use by other applications; discover and load MCP tools dynamically; maintain type safety across the MCP boundary. The goal: Orchestral tools work natively across both the Orchestral ecosystem and the broader MCP ecosystem, without manual translation.

\subsubsection{Lightweight Deployment Model}

Unlike IDE-centric frameworks requiring Electron runtimes and editor integration, or systems requiring FastAPI servers and PostgreSQL databases, Orchestral is designed for embedding. The framework is a single Python package with minimal dependencies, enabling deployment scenarios impractical for heavier alternatives:

\begin{itemize}
    \item \textbf{Research projects:} Include Orchestral as a dependency, not a platform. Researchers control the architecture; Orchestral provides capabilities.
    \item \textbf{Low-resource servers:} Deploy on budget VPS instances, edge devices, Raspberry Pis. No database server, no message queue, no container orchestration required.
    \item \textbf{Serverless functions:} Cold-start friendly with no persistent infrastructure. Deploy agents as AWS Lambda, Google Cloud Functions, or Cloudflare Workers.
    \item \textbf{Docker containers:} Minimal image size (base Python + pip), fast builds, efficient resource usage.
    \item \textbf{Embedded in applications:} Add agent capabilities to existing Python applications without architectural commitment. Orchestral adapts to your codebase, not vice versa.
\end{itemize}

This deployment model contrasts sharply with frameworks that become your application's foundation. Orchestral's footprint is a Python interpreter and pip, nothing more. For researchers deploying agents as part of larger projects, or developers adding AI capabilities to existing systems, this lightweight approach eliminates architectural friction.

\section{Conclusion}

Orchestral demonstrates that careful architectural design can provide both simplicity and power in AI agent frameworks. By prioritizing synchronous execution, provider agnosticism, and type safety through Python's native features, the framework reduces many sources of complexity present in existing solutions.

For scientific computing researchers, Orchestral combines production robustness (cost tracking, security, error handling) with research affordances (LaTeX integration, streaming with interrupts, reproducible workflows). The framework provides both exploration and deployment capabilities through architectural clarity that makes every operation understandable.

The modular architecture with strict separation of concerns enables extensibility without complexity. Adding providers, tools, or hooks requires implementing clear interfaces without navigating sprawling codebases. The framework's independence from application-specific concerns enables reuse across contexts: CLI tools, web applications, notebooks, API services, or embedded in larger systems.

We believe AI agent frameworks should be:
\begin{itemize}
    \item \textbf{Understandable:} Control flow should be explicit, not hidden in async machinery
    \item \textbf{Debuggable:} Errors should be traceable, not lost in event loops
    \item \textbf{Portable:} Work should not be locked to a single vendor
    \item \textbf{Type-safe:} Schemas should derive from code, not manual JSON
    \item \textbf{Production-ready:} Features like persistence and cost tracking should be built-in
    \item \textbf{Research-friendly:} LaTeX export and reproducibility should be first-class
\end{itemize}

Orchestral pursues these goals through architectural discipline rather than feature accumulation. The result is a framework that researchers can understand completely, developers can debug straightforwardly, and both can deploy confidently, all in a single lightweight Python codebase.

Orchestral is available for use in research and production applications.

\section*{Acknowledgments}

We are especially grateful to Ioannis Michaloliakos, Marco Knipfer, Tony Menzo, and Emilie Panek for their valuable contributions to testing and refining the Orchestral framework. Their detailed feedback and engagement with the framework during development helped shape its design and capabilities.
We thank Konstantin Matchev, Katia Matcheva, Leonardo Pagliaro and Gaurav Shukla, for their input, testing, and feedback during development.

We acknowledge helpful discussions and encouragement from Sergei Gleyzer at the University of Alabama, and thank our collaborators at Fermilab: Bogdan Dobrescu, George T. Fleming, Stefan Höche, Stephen Mrenna, and Prasanth Shyamsundar for their contributions.

\printbibliography

\end{document}